\documentclass{article}

\usepackage[english]{babel}

\usepackage[letterpaper,top=2cm,bottom=2cm,left=3cm,right=3cm,marginparwidth=1.75cm]{geometry}

\usepackage{amsmath, amsfonts, amssymb}
\usepackage{graphicx}
\usepackage[colorlinks=true, allcolors=blue]{hyperref}

\usepackage[backend=biber,style=numeric,sorting=none]{biblatex}
\usepackage{newunicodechar}
\newunicodechar{₃}{$_3$}
\usepackage{booktabs}
\usepackage{multirow}
\usepackage{adjustbox}

\makeatletter
\renewcommand{\maketitle}{
    \begin{center}
        \includegraphics[width=1\textwidth]{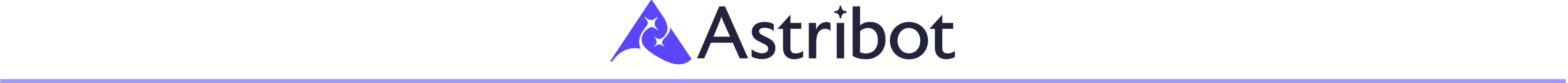}\\[1.5cm] 
        {\LARGE \bfseries \@title}\\[0.5cm]
        {\large \@author}\\[0.5cm]
        {\large \@date}
    \end{center}
}
\makeatother

\title{Asynchronous Fast-Slow Vision-Language-Action Policies for Whole-Body Robotic Manipulation}
\author{Astribot Team \\ \small astribot\_ai@astribot.com \\
  {\small Full Author List in \hyperref[sec:contrib]{Contributions and Acknowledgments}}}
  
  \date{}

\begin{document}

\maketitle

\begin{abstract}
Most Vision-Language-Action (VLA) systems integrate a Vision-Language Model (VLM) for semantic reasoning with an action expert generating continuous action signals, yet both typically run at a single unified frequency. As a result, policy performance is constrained by the low inference speed of large VLMs. This mandatory synchronous execution severely limits control stability and real-time performance in whole-body robotic manipulation, which involves more joints, larger motion spaces, and dynamically changing views. We introduce a truly asynchronous Fast-Slow VLA framework (DuoCore-FS), organizing the system into a fast pathway for high-frequency action generation and a slow pathway for rich VLM reasoning. The system is characterized by two key features. First, a latent representation buffer bridges the slow and fast systems. It stores instruction semantics and action-reasoning representation aligned with the scene-instruction context, providing high-level guidance to the fast pathway. Second, a whole-body action tokenizer provides a compact, unified representation of whole-body actions. Importantly, the VLM and action expert are still jointly trained end-to-end, preserving unified policy learning while enabling asynchronous execution. DuoCore-FS supports a 3B-parameter VLM while achieving 30 Hz whole-body action-chunk generation, approximately three times as fast as prior VLA models with comparable model sizes. Real-world whole-body manipulation experiments demonstrate improved task success rates and significantly enhanced responsiveness compared to synchronous Fast-Slow VLA baselines. The implementation of DuoCore-FS, including training, inference, and deployment, is provided to commercial users by Astribot as part of the Astribot robotic platform.

\end{abstract}

\section{Introduction}
Vision–language–action (VLA) \cite{brohan2023-rt1, octomodelteam2024-octo, walke2024-bridgedatav2} models have recently attracted substantial attention in general-purpose robotic manipulation, as they provide a unified framework for learning policy that takes visual observations and linguistic instructions as inputs and generates corresponding action poses as outputs. To mitigate the limitations caused by the scarcity of robotic data, some works \cite{pmlr-RT2, kim2024-openvla} directly leverage VLMs pretrained on internet-scale image–text corpora, extending the auto-regressive next-token prediction paradigm to generate action tokens. This strategy transfers pretrained cross-modal knowledge to robot policies, aligning action outputs with visual and linguistic signals. 

While autoregressive VLA methods have shown promise, the sequential, token-by-token generation of discrete action tokens often suffers from low inference speed and limited  action accuracy. To enable the generation of high-frequency continuous action chunks, a dual-system VLA architecture has been proposed in several works \cite{black2410-pi0, nvidia2025-gr00t}, inspired by human cognitive processing \cite{bonnefon2020-fastslow}. The dual-system VLA architecture jointly fine-tunes a VLM and a lightweight diffusion-based action expert, where the VLM acts as a slow, deliberative reasoning module and the action expert provides fast, reactive control. 

In most dual-system VLA implementations, the VLM and action expert still operate synchronously, so the fast module’s execution frequency is ultimately constrained by the slow VLM’s inference speed. Especially, this limitation becomes increasingly severe as a recent trend has emerged in which VLA systems \cite{kim2024-openvla, dang2025-rynnec, intelligence2025-pi06, generalist2025-gen0} adopt much larger VLM backbones, further widening the gap between reasoning latency and the control-rate requirements of real-world manipulation. To address this bottleneck, several recent works \cite{figure2025-helix, song2025-hume, chen2025-fastinslow} introduce asynchronous dual-system VLA, allowing the two subsystems to run at independent frequencies rather than being locked to the VLM’s slow update rate. In this formulation, the slow system module performs infrequent high-level deliberation, while the fast system module updates at a much higher frequency to generate low-level actions for real-time control, as previously described in \cite{cui2025-openhelix}.

Our approach differs from existing asynchronous dual-system frameworks such as FiS-VLA \cite{ chen2025-fastinslow}, OpenHelix \cite{cui2025-openhelix}, Helix \cite{figure2025-helix}, and Hume \cite{song2025-hume} in how the slow and fast subsystems interact during inference or training. In our design, the slow system and fast system run fully in parallel: the slow module engages in multi-faceted reasoning at a low frequency, e.g., generating discrete action tokens and providing high-level latent representations to guide the fast system, while the fast module translates the latest latent representations from the slow module together with real-time visual observations and proprioceptive states to generate high-frequency continuous actions. In contrast, FiS-VLA and OpenHelix do not perform parallel inference; instead, they enforce a fixed frequency ratio between the slow and fast modules, causing the fast controller’s update cycle to depend on predetermined scheduling rather than true asynchrony. Our method adopts a parallel slow–fast design akin to Helix. However, Helix is not open-sourced, and publicly available materials indicate that its slow module outputs only latent representations, without describing any mechanism for generating reasoning or action tokens at inference time. Meanwhile, Hume instead employs a cascaded slow–fast architecture that, although capable of asynchronous parallel inference and with the slow module able to generate action tokens, is not trained end-to-end. This cascaded architecture restricts the interaction between high-level reasoning and real-time control, as the lack of end-to-end training prevents the two components from being jointly optimized.

In addition, whole-body robotic manipulation requires coordinating a larger number of joints within an expanded motion space, while adapting to dynamic changes in camera viewpoints and environmental context, such as tasks that may involve torso rotation, bending, or reaching across obstacles. These factors not only increase the dimensionality of the control problem compared to dual-arm manipulation, but also intensify the need for VLA policies that can produce precise, high-frequency action signals across the entire body.

Motivated by these insights, we introduce DuoCore-FS, a fast–slow VLA framework that achieves high-frequency whole-body action generation while leveraging the reasoning capacity of a large VLM. The system consists of a slow pathway for multi-modal deliberation and a fast pathway for continuous control, connected through a latent representation buffer that conveys both semantic information from the VLM and high-level action reasoning. Key aspects of DuoCore-FS include:

\begin{itemize}
  \item \textbf{Truly parallel and asynchronous fast–slow execution:} A high-frequency fast pathway generates continuous whole-body actions for real-time control, while a truly asynchronous low-frequency slow pathway performs semantic understanding and high-level reasoning. Benefiting from this design, the overall action-chunk generation frequency is ultimately governed by the fast pathway.
  \item \textbf{A bridge buffer enriched by the alignment across V-L-A:} The slow system updates the buffer with latent representations, which are transmitted via a bridging mechanism to the fast system, allowing the fast pathway to integrate semantic cues from vision and language with high-level guidance for high-frequency continuous whole-body control.
  \item \textbf {Whole-body action tokenizer:} Provides a compact, unified representation that captures the structured patterns of high-dimensional whole-body joint configurations, enabling coordinated control across the full robot body.
  \item \textbf {End-to-end training:} Both pathways are trained simultaneously, allowing the slow and fast modules to be optimized jointly, promoting tight semantic–control alignment without relying on separate or hand-engineered modules.
\end{itemize}

\section{Related Work}
\subsection{Vision Language Action Policies}
Vision--Language--Action (VLA) policies unify visual perception, language understanding, and action generation within a single model, enabling end-to-end instruction-conditioned robotic manipulation. Early autoregressive approaches~\cite{brohan2023-rt1,pmlr-RT2,rdt2,kim2024-openvla,zhao2025cot,pertsch2025fast,zheng2025universal,cen2025worldvla} extend large vision--language models to output action tokens for cross-modal control, but their sequential token-by-token decoding inherently limits control frequency. Diffusion-based VLA methods~\cite{black2410-pi0,intelligence2025pi_,zhai2025igniting,liu2024rdt,bjorck2025gr00t,xu2025a0,qu2025eo} mitigate this issue by generating continuous action chunks in parallel, yielding smoother and more stable low-level control. Reinforcement-driven VLA systems~\cite{tan2025interactive,lu2025vla,chen2025pi_,amin2025pi} further improve long-horizon reasoning by optimizing policies through reward signals.

Despite methodological differences, most existing VLA frameworks still follow a single-stream design, coupling visual–language inference and action generation within the same computation pipeline. As VLMs continue to grow in size and computational complexity~\cite{kim2024-openvla, dang2025-rynnec, intelligence2025-pi06, generalist2025-gen0}, their inference speed increasingly fails to meet real-time control requirements, and lightweight action heads cannot overcome this bottleneck. These limitations have motivated the emergence of fast–slow VLA architectures, where a low-frequency slow subsystem handles semantic reasoning while an independent fast subsystem executes high-frequency actions. Following this direction, we propose a parallel fast--slow design for whole-body manipulation, enabling large VLMs to provide semantic guidance without compromising control frequency.

\subsection{Dual-System VLA Architectures}
Dual-system VLA architectures are inspired by the fast--slow cognition theory~\cite{bonnefon2020-fastslow}, typically employing a large VLM as the low-frequency ``slow system'' and a lightweight diffusion-based controller as the high-frequency ``fast system.'' Several recent methods nominally follow this paradigm, including DP-VLA~\cite{han2024dual}, HiRT~\cite{zhu2025scaling}, Robodual~\cite{bu2024towards}, G0-VLA~\cite{jiang2025galaxea}, and Pi05~\cite{intelligence2025pi_}: the slow system outputs task semantics, feature representations, or latent goals, while the fast system generates executable actions conditioned on these high-level signals. However, in all these approaches the two subsystems still operate synchronously, requiring the fast system to wait for updates from the slow system. As a result, the overall action-generation frequency is fundamentally constrained by the slow inference speed of the VLM, preventing true fast--slow fully asynchronous execution.

In contrast, FiS-VLA~\cite{chen2025-fastinslow} and OpenHelix~\cite{cui2025-openhelix} introduce asynchronous execution, but the interaction between the slow and fast subsystems is still governed by a fixed scheduling ratio (e.g., one slow update for every several fast steps). As a result, the fast subsystem cannot operate truly independently; its update cycle remains indirectly dictated by the predetermined schedule rather than by real-time perception or system dynamics. Helix~\cite{figure2025-helix} and LCB~\cite{shentu2024llms} propose designs that are conceptually closer to a fully parallel slow--fast architecture, but neither provides open-source implementation details. For Helix, the official materials~\cite{figure2025-helix} do not offer a detailed description of how the slow subsystem performs explicit reasoning or action generation during inference. LCB~\cite{shentu2024llms} positions the slow subsystem primarily as a conversational interface that provides language-level guidance, and the paper does not report real-world robotic evaluations. By comparison, our framework employs a slow subsystem that produces action-level guidance and is validated on real-world manipulation tasks, while maintaining a fully parallel and asynchronous fast–slow architecture and supporting end-to-end training of both pathways.

\section{Methodology}
We introduce a fast--slow Vision-Language-Action (VLA) policy framework designed to couple high-level semantic reasoning with real-time whole-body control. Complex manipulation tasks require both abstract interpretation of visual–linguistic instructions and rapid motor responses, yet these demands naturally operate at different temporal scales. To bridge this gap, our framework leverages a large Vision-Language Model (VLM) to provide temporally sparse but semantically rich guidance, while a lightweight diffusion policy continuously generates actions. This separation of semantic inference and control enables the system to maintain strong task-level understanding without compromising reactivity, allowing robots to perform semantically grounded whole-body manipulation in dynamic and open-world scenarios.

\subsection{Architecture}
The proposed VLA policy consists of two asynchronously operating subsystems: a low-frequency slow system for semantic reasoning and intent extraction, and a high-frequency fast system for real-time whole-body control, as illustrated in Figure~\ref{fig:framework}. The slow system uses a large Vision-Language Model (VLM) to generate structured semantic hidden states from visual observations and task prompts, which are periodically refreshed into a bridge buffer. The fast system fetches the latest latent representation from this buffer at control-rate frequency, fuses them with current perceptual inputs, and produces continuous actions through a Transformer-based diffusion-policy decoder.

\begin{figure}[!t]
\centering
\includegraphics[width=0.95\linewidth]{framework.png}
\caption{\label{fig:framework}Overview of the proposed fast--slow asynchronous VLA policy framework. The slow system (bottom) operates at a low frequency of 1--3 Hz, where a large Vision-Language Model (VLM) parses task instructions, visual observations, and proprioceptive states to produce high-level semantic hidden states, including text embeddings, reasoning features, and learnable fusion queries. These representations are periodically written into a bridge buffer. The fast system (top) fetches the latest latent semantic representations from the bridge buffer at a high control rate of 25--30 Hz, integrates them with current visual features and proprioceptive states, and uses a Transformer-based diffusion-policy decoder to generate smooth, continuous, and fully coordinated whole-body actions.}
\end{figure}

\paragraph{Slow system: semantic and reasoning representations construction.}
The slow system operates at a low update frequency and performs high-level semantic reasoning using a compact yet expressive Vision-Language Model (VLM). In practice, this component can be instantiated with models such as PaliGemma-3B~\cite{beyer2024paligemma}, Qwen2.5-VL-3B, or Qwen2.5-VL-7B~\cite{bai2025qwen2}, depending on the computational budget and task requirements. Given the current multimodal observation
\[
o_t = \{I^1_t, \ldots, I^n_t, q_t\},
\]
where \(I^i_t\) denotes the visual image captured by the \(i\)-th camera at time \(t\), and \(q_t\) represents the proprioceptive state, together with the task prompt \(l\), the VLM adaptively generates semantic outputs \(r_t\) that best support the manipulation task. These outputs may include chain-of-thought reasoning traces, bounding-box predictions, subtask indicators, and coarse discrete action tokens. The coarse action tokens encode high-level manipulation intent in a compact and task-aligned form, providing strong priors for downstream whole-body control. The fast system leverages these priors to generate fine-grained and continuous trajectories.
The VLM is trained using a negative log-likelihood objective parameterized by \(\phi\), where \(\phi\) denotes all learnable parameters of the slow VLM subsystem:
\[
\mathcal{L}_{\text{slow}}(\phi)
=
\mathbb{E}_{(o_t,\, l,\, r_t)\sim\mathcal{D}}
\big[
- \log p(r_t \mid o_t,\, l)
\big],
\]
which encourages the model to produce semantic and reasoning representations that are aligned with the multimodal context defined by vision, proprioception, and task instructions.

\paragraph{Bridge buffer: asynchronous semantic and reasoning representation caching.}
To enable asynchronous information flow between the slow and fast systems, we introduce a bridge buffer $\mathcal{B}_t$, which stores task-relevant semantic and reasoning representations produced by the VLM. Unlike synchronous policies where high-level semantics must be updated at every step, the bridge buffer decouples the generation and consumption of representations, allowing the system to operate at different frequencies: the slow system refreshes $\mathcal{B}_t$ at a low rate of 1--3 Hz, while the fast system fetch the latest representations at 25--30 Hz to condition whole-body action generation. In practice, $\mathcal{B}_t$ stores both the instruction embeddings and the fusion-query embeddings, and serves as a differentiable interface through which the slow system, the fast system, and the fusion queries are trained end-to-end.

To populate the bridge buffer, the slow system receives learnable fusion queries $q_\psi$ that aggregate task-relevant semantic cues and reasoning information. These queries attend across visual observations, proprioceptive states, task instructions, and the VLM’s internal semantic space, producing structured, manipulation-oriented representations. These representations are then written into $\mathcal{B}_t$ and remain valid across multiple fast-system control cycles. Importantly, the parameters $\psi$ of the fusion queries are trained through the fast system's action-generation objective, enabling them to progressively learn semantic and reasoning features that best support whole-body action generation.

Furthermore, although the semantic aggregation produced by the fusion queries provides high-level conditioning signals for action generation, we still feed the original task instruction representation into the fast system. This is because the task instruction remains constant throughout execution, whereas relying solely on the task-related semantic information from the fusion queries may result in the loss of crucial task details or cause the representation to overly focus on patterns directly related to action generation. By supplying both the original instruction and the semantic representations stored in the buffer, the fast system is able to maintain stable linguistic constraints while avoiding excessive reliance on dynamically extracted reasoning, thereby further improving overall task performance and long-term stability.

\paragraph{Fast system: diffusion-policy whole-body action generator.}
The fast system operates at approximately 25–30 Hz and is responsible for generating continuous whole-body actions conditioned on guidance from the slow system. At each action step $t$, the fast system functions as a Pi0-small\cite{black2410-pi0} style diffusion-policy network, modeling the action distribution based on current visual observations, proprioceptive states, the latest fusion queries stored in the bridge buffer, and raw instruction embeddings. All conditioning modalities are projected into a shared embedding space and concatenated into a unified token sequence. A multi-layer transformer encoder processes this sequence and generates the conditioning signal for diffusion-policy action generation.

Following the continuous-time formulation of diffusion policies, the fast system learns a conditional vector field over action chunks. Given an expert action chunk $A_{t:t+T-1}$, we sample Gaussian noise $\varepsilon\sim\mathcal{N}(0,I)$ and a diffusion time $\tau\in(0,1]$, and construct noisy actions as
\[
X_\tau = \tau\,\varepsilon + (1-\tau)\,A_{t:t+T-1}.
\]

The fast system is parameterized by $\theta$ and is optimized to predict a denoising vector field. Formally, the denoising direction is given by $v_\theta(X_\tau, o_t, \mathcal{B}_{t-\Delta_t})$, where $\Delta_t$ denotes the time-varying observation delay between the fast and slow systems. The optimization objective is:
\[
\mathcal{L}_{\text{fast}}(\theta, \psi) =
\mathbb{E}_{A, \varepsilon, \tau, \Delta_t}
\Bigl[
\bigl\|\,\varepsilon - A_{t:t+T-1}
      - v_\theta(X_\tau, o_t, \mathcal{B}_{t-\Delta_t})\,\bigr\|_2^2
\Bigr].
\]

\subsection{Action Tokenizer for Whole-Body Manipulation}
The Astribot whole-body platform exposes a structured and fully controllable 25\,DoF action space. To improve spatial generalization across different workspaces, we adopt Cartesian delta pose control for all major kinematic units. The torso is commanded by a nine-dimensional delta pose vector consisting of a three-dimensional translational offset and a six-dimensional continuous $\mathrm{SO}(3)$ orientation representation, and the left and right arms follow the same parameterization. The grippers, in contrast, are controlled using absolute opening-width commands. To obtain a compact yet expressive discrete representation of whole-body actions, we further build a geometry-aware action tokenizer inspired by the residual VQ-VAE design used in RDT2\cite{rdt2}. At each action step $t$, the action $a_t \in \mathbb{R}^{29}$ is factorized into three semantically grounded components,
\[
a_t = [a_t^{\text{pos}},\, a_t^{\text{rot}},\, a_t^{\text{grip}}],
\]
where $a_t^{\text{pos}}$ denotes the Cartesian position deltas of the torso and end-effectors, $a_t^{\text{rot}}$ represents their orientation deltas using a continuous $\mathrm{SO}(3)$ parameterization, and $a_t^{\text{grip}}$ corresponds to the \emph{absolute} opening commands of the left and right grippers.

During action tokenization, an action chunk is defined as $\mathbf{A}_{t:t+T-1}$, representing the short-horizon temporal context of actions around the current step. Each chunk is decomposed into structured position, orientation, and gripper streams:
\[
P \in \mathbb{R}^{T \times 9}, \qquad
R \in \mathbb{R}^{T \times 18}, \qquad
G \in \mathbb{R}^{T \times 2}.
\]
Each control stream is first processed by an independent lightweight 1D convolutional encoder--decoder, producing latent features
$Z^{\text{pos}}, Z^{\text{rot}}, Z^{\text{grip}}$. These latent sequences are then discretized using stream-specific residual vector quantization (RVQ)~\cite{lee2022autoregressive} modules. Each RVQ stack maintains its own codebook with a size of 1024.

During training, each RVQ-VAE is optimized to reconstruct its corresponding action stream. For the position and gripper branches, we use an $\ell_2$ reconstruction loss:
\[
\mathcal{L}^{\text{pos}}_{\text{rec}}=\|\hat{P}-P\|_2^2, \qquad
\mathcal{L}^{\text{grip}}_{\text{rec}}=\|\hat{G}-G\|_2^2 .
\]

For the rotation branch, each 6D orientation vector is first converted into a rotation matrix, and the reconstruction loss is computed using the geodesic distance on $\mathrm{SO}(3)$:
\[
\mathcal{L}^{\text{rot}}_{\text{rec}}
=\frac{1}{3}\left(
\ell_{\mathrm{geo}}(\hat{R}_{\text{torso}},R_{\text{torso}})
+\ell_{\mathrm{geo}}(\hat{R}_{\text{left}},R_{\text{left}})
+\ell_{\mathrm{geo}}(\hat{R}_{\text{right}},R_{\text{right}})
\right),
\]
where $\ell_{\mathrm{geo}}$ denotes the $\mathrm{SO}(3)$ geodesic loss.

Each residual quantizer additionally incurs a standard VQ objective, including a commitment loss and a codebook loss. We denote the sum over all quantizers as $\mathcal{L}_{\mathrm{vq}}$. The overall training objective is
\[
\mathcal{L}_{\text{token}}
=
\mathcal{L}^{\text{pos}}_{\text{rec}}
+
\mathcal{L}^{\text{rot}}_{\text{rec}}
+
\mathcal{L}^{\text{grip}}_{\text{rec}}
+
\mathcal{L}_{\mathrm{vq}}.
\]

After training, the RVQ-VAE encoders serve to convert continuous action chunks into discrete token sequences.

\subsection{Cross-Timescale Co-Training}
Our training procedure is organized into two stages to ensure effective end-to-end optimization across different temporal scales. In the first stage, the slow system is trained independently using standard vision–language modeling objectives. This phase focuses on learning high-level reasoning from visual observations and task prompts, while simultaneously aligning visual, linguistic, and action-related cues within the VLM so that all modalities are embedded in a coherent semantic space. 

In the second stage, we jointly train the fast and slow systems to enable end-to-end optimization across different temporal scales. A key challenge is that the slow system operates at only 1--3\,Hz, while the fast controller runs at 25--30\,Hz. If both systems were updated using the same observation at each training step, this would lead to a significant train--inference mismatch. To address this issue, we introduce a cross-timescale sampling strategy that explicitly simulates the asynchronous behavior of the deployed system.

To accurately reproduce the asynchronous timing characteristics observed in real deployment, we introduce a cross-timescale sampling strategy during joint training. Given the slow-system input observation $o_{t_0}$, the fast system receives a temporally shifted observation $o_{t_0+\Delta}$, where
\[
\Delta \sim \mathcal{U}[0,\,\Delta_{\max}].
\]
The upper bound $\Delta_{\max}$ represents the maximum observation delay that the fast system may experience relative to slow-system inference. Let $T_{\text{slow}}$ and $T_{\text{fast}}$ denote the update periods of the slow and fast systems, respectively, with $f_{\text{cam}}$ denoting the camera frame rate. The maximal temporal offset is approximated as
\[
\Delta t_{\max} \approx 
T_{\text{slow}}
+
\Big\lfloor T_{\text{slow}}/{T_{\text{fast}}} \Big\rfloor
T_{\text{fast}},
\]
which converts to a frame-level delay
\[
\Delta_{\max} \approx 
\big\lfloor \Delta t_{\max} \, f_{\text{cam}} \big\rfloor.
\]

\begin{figure}
\centering
\includegraphics[width=0.95\linewidth]{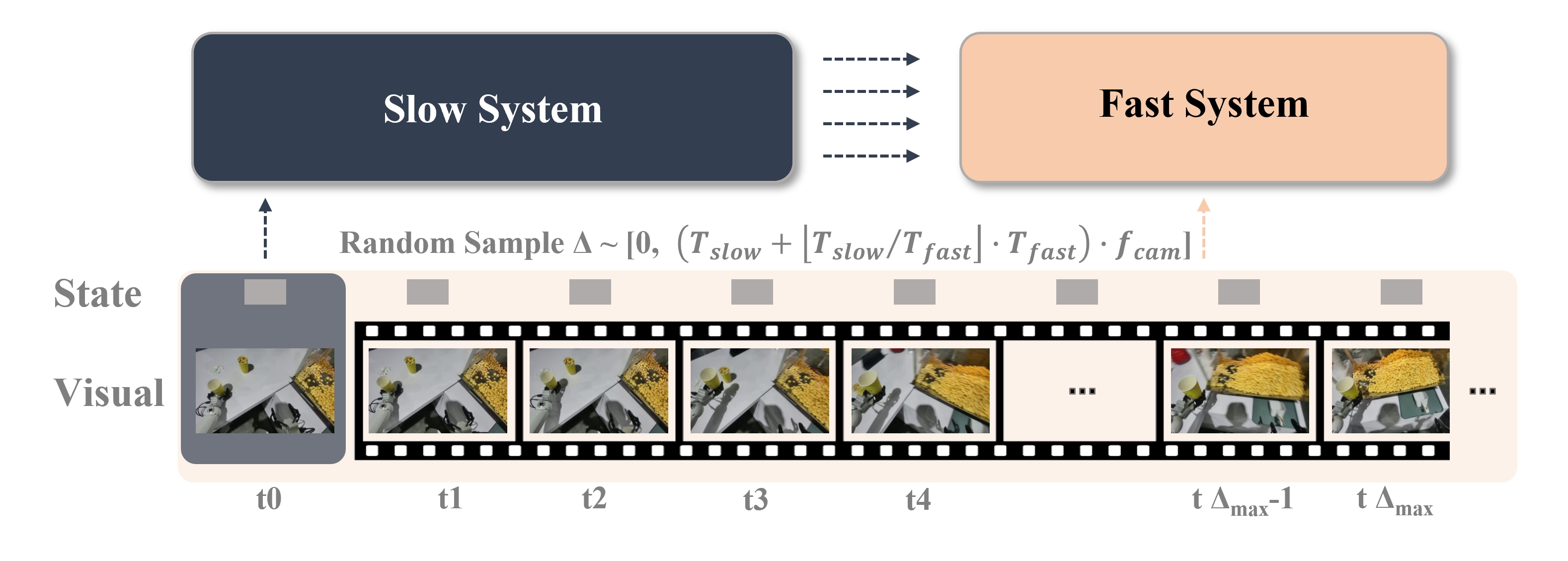}
\caption{\label{fig:co_training}Cross-timescale co-training. The slow system processes observation $o_{t_0}$, while the fast system receives a shifted observation $o_{t_0+\Delta}$ with $\Delta \sim \mathcal{U}[0,\,\Delta_{\max}]$. This sampling strategy mimics 
the asynchronous timing during real deployment, enabling consistent end-to-end optimization.}
\end{figure}

This sampling mechanism trains the fast system under the same asynchronous observation–conditioning pattern as in inference. As illustrated in Fig.~\ref{fig:co_training}, the slow system updates the semantic buffer only at sparse intervals, while the fast system acts on observations with temporal delay. This cross-timescale simulation removes the train–inference mismatch caused by their different operating frequencies.

In this stage, the Fast--Slow VLA framework is optimized jointly under the cross-timescale sampling scheme. The unified optimization objective is
\[
\mathcal{L}_{\text{joint}}(\phi, \theta, \psi)
=
\lambda_{\text{slow}}\,\mathcal{L}_{\text{slow}}
+
\lambda_{\text{fast}}\,\mathcal{L}_{\text{fast}},
\]
where we set $\lambda_{\text{slow}}=1$ and $\lambda_{\text{fast}}=10$. This weighting indicates that the slow system has been pretrained in Stage~1, while the fast controller (together with the bridge tokens) requires stronger supervision to master precise whole-body action under asynchronous semantic guidance.

\subsection{Asynchronous Inference Pipeline for Real-Time Control}
During real-world deployment, the proposed fast–slow VLA policy operates in a fully asynchronous manner, as illustrated in Figure~\ref{fig:framework}. The slow system runs at a low frequency of 1--3,Hz, where a large Vision-Language Model (VLM) processes the current visual observation and task instruction to generate text representations, reasoning representations, and bridge embeddings that aggregate task-critical information. These representations are then periodically written into a bridge buffer, which serves as a decoupling interface between the two subsystems.

To further reduce the inference latency of the slow system, we adopt a Jacobi-style parallel decoding strategy inspired by Consistency Large Language Models (CLLMs)~\cite{kou2024cllms}. This approach accelerates the VLM’s reasoning process by enabling multiple iterative refinements to be computed in parallel, thereby lowering the effective update cost of the slow system without altering its semantic capacity.

The fast system executes at a high action frequency of 25--30\,Hz and continuously fetches the most recent latent representation from the buffer. At each step, it integrates these representations with current visual features and proprioceptive states, forming a unified conditioning input for the Transformer-based diffusion-policy. Even if the slow system has not yet produced new reasoning outputs, the fast system operates smoothly using the previously refreshed instruction and bridge embeddings. This asynchronous design ensures stable, real-time whole-body control while preserving alignment with the high-level intent inferred by the slow system, enabling the robot to react promptly to environmental changes without being constrained by the slower VLM update cycle.

\section{Experiments}
\subsection{Experiment Settings}
\noindent\textbf{Robotic Platform.} To validate our fast–slow VLA system, we conduct experiments on the Astribot S1 platform. Astribot S1 is a mobile dual-arm manipulator featuring two 7-DoF arms with parallel-jaw grippers, a 4-DoF articulated torso, a 2-DoF head, and a 3-DoF omnidirectional mobile base, as illustrated in Figure~\ref{fig:framework}. Demonstration data are collected through teleoperation, providing high-quality trajectories for training fast–slow VLA policies. Detailed descriptions of the hardware, teleoperation interface, and algorithm deployment workflow are provided in Astribot Suite\cite{gao2025-astribot}.

\begin{figure}
\centering
\includegraphics[width=0.95\linewidth]{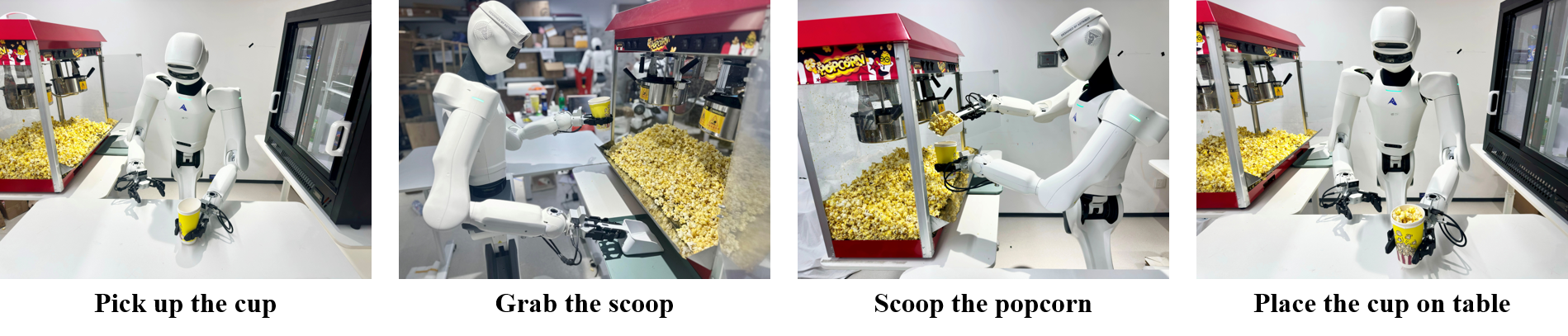}
\caption{\label{fig:popcorn_pipeline}\textbf{Popcorn-scooping task pipeline.} }
\end{figure}

\noindent\textbf{Task and Dataset.} 
We collected 1,780 demonstration trajectories (10.22 hours in total) in a commercial popcorn kiosk scenario for training. The dataset contains a long-horizon popcorn-scooping task and a short-horizon task of closing the beverage cabinet door. The popcorn-scooping task is a whole-body manipulation problem that requires coordinated control of both arms and the torso to complete a sequence of sub-tasks, as illustrated in Fig.~\ref{fig:popcorn_pipeline}. We further decompose the long-horizon popcorn-scooping task into four sub-tasks:

\begin{itemize}
\item [1)] 
\textbf{Pick up the cup:} The robot reaches toward and grasps an empty cup placed on the serving table.
\item [2)] 
\textbf{Grab the scoop:} The robot rotates its upper body while carrying the cup from the serving area to the popcorn machine, then grasps the scoop.
\item [3)] 
\textbf{Scoop the popcorn:} The robot scoops popcorn with its right arm and pours it into the cup.
\item [4)] 
\textbf{Place the cup on table:} The robot turns back to the serving area and places the filled cup back onto the table.
\end{itemize}

\noindent\textbf{Evaluation Metric.} 
Each episode consists of four sequential manipulation subtasks illustrated in Figure~\ref{fig:popcorn_pipeline}, executed under a single high-level instruction (e.g., “pick up the paper cup and scoop popcorn”). Because the subtasks must be completed in order, later steps become unreachable if earlier ones fail. This setting reflects realistic long-horizon manipulation, where errors propagate and early failures prevent further progress. We report per-subtask conditional success rates, measuring success only over trials in which a subtask becomes reachable. This isolates the performance of each stage while respecting the task’s sequential structure. We also report overall task success, defined as completing all four subtasks end-to-end.

\noindent\textbf{Training Details.} 
For our DuoCore-FS, we adopt a flexible fast--slow dual-system architecture, where the slow system can be instantiated with different VLM backbones. In principle, any vision-language(-action) model can be plugged into the slow system. In our experiments, we instantiate the slow system with open-sourced $\pi_{0}$-FAST~\cite{pertsch2025fast}, which is built on the 3B PaliGemma~\cite{beyer2024paligemma} backbone and provides pre-trained weights for visuomotor control. Training uses images from three viewpoints: the head camera, the left-hand camera, and the right-hand camera. All input images for both the fast and slow systems are resized to $224 \times 224$. The frame-level delay between the two systems is randomly sampled as $\Delta \sim \mathcal{U}[0, 25]$, and the action chunk length is set to $T = 32$. In the first stage, we independently train the slow system to autoregressively predict action tokens encoded by the RVQ-VAE for 30 epochs. Training is conducted on 24 NVIDIA H100 GPUs (batch size 25 per GPU). We adopt a cosine-decay learning-rate schedule with a peak learning rate of $1\times10^{-4}$ and a minimum learning rate of $3\times10^{-6}$. In the second stage, we jointly train the fast and slow systems for 12 epochs using a cosine schedule with a peak learning rate of $5\times10^{-5}$ and a minimum learning rate of $3\times10^{-6}$. For all remaining methods, we follow Stage-2–like configurations. All models are trained in BF16.

\noindent\textbf{Inference Details.} 
We evaluate the inference performance of DuoCore-FS and $\pi_{0}$\cite{black2410-pi0} on the same compute platform, with all experiments conducted on a single NVIDIA RTX 4090 GPU. For $\pi_{0}$\cite{black2410-pi0}, we use the official JAX implementation for inference, while all other models are executed in PyTorch. The fast system in our method is compiled to TensorRT in BF16.

\begin{figure}
\centering
\includegraphics[width=0.80\linewidth]{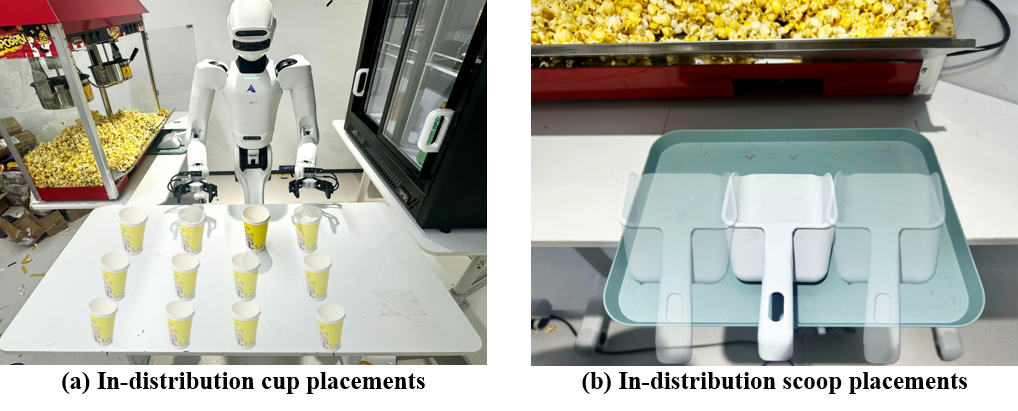}
\caption{\label{fig:cup_scoop_indistribution}\textbf{In-distribution benchmark setup.} The cup and the scoop are placed at diverse locations. The figure provides a schematic overview and does not enumerate all test positions.}
\end{figure}

\subsection{Experimental Results}
\noindent\textbf{In-Distribution Results.} To fully evaluate different methods on the popcorn-scooping task, we position the cup and scoop at various in-distribution locations, as Figure~\ref{fig:cup_scoop_indistribution} illustrates.
As Table~\ref{tab:result_in_distribution} shows, our slow system (DuoCore-FS-slow) has a lower success rate and runs more slowly than $\pi_{0}$~\cite{black2410-pi0}. This is consistent with expectations: the reasoning action tokens serve as a lossy abstraction of continuous actions, which hinders fine-grained behaviors such as accurately aligning with the scoop handle. Moreover, these tokens are generated autoregressively, making them inherently slower to produce. In contrast, DuoCore-FS achieves success rates comparable to $\pi_{0}$~\cite{black2410-pi0} while running at nearly $3\times$ its inference speed. This demonstrates the effectiveness of our dual system architecture: the slow system’s reasoning tokens provide effective guidance to the fast system, boosting success, while asynchronous inference delivers the speedup.

\begin{table}[htbp]
\caption{\label{tab:result_in_distribution}\textbf{Results in-distribution.} Success is assessed by humans based on task completion. For example, in the "Scoop the popcorn" task, the robot must scoop popcorn with the scoop and accurately pour it into the cup. Since we evaluate under coarse instructions, all subtask success rates are reported as conditional success rates. DuoCore-FS-slow is our slow system.}
\centering
\resizebox{\linewidth}{!}{%
\begin{tabular}{lccccc c}
\toprule
Method & Pick up the cup & Grab the scoop & Scoop the popcorn & Place the cup on table & Overall S.R. $\uparrow$ & Inference Frequency (Hz) \\
\midrule
$\pi_{0}$~\cite{black2410-pi0} & 95\% (19/20) & 89.5\% (17/19) & 100\% (17/17) & 100\% (17/17) & 85\% (17/20) & 12.5 \\
DuoCore-FS-slow            & 100\% (20/20) & 75\% (15/20)  & 73.3\% (11/15) & 100\% (11/11) & 55\% (11/20) & 3.27 \\
DuoCore-FS                 & 100\% (20/20) & 90\% (18/20)  & 100\% (18/18) & 100\% (18/18) & 90\% (18/20) & \textbf{32.3} \\
\bottomrule
\end{tabular}%
}
\end{table}

\noindent\textbf{Out-of-Distribution Results.} To evaluate generalization, we place the cup at unseen locations such as near the table edge and test whether the model can still complete the task. Table.~\ref{tab:result_out_distribution} indicates that our slow system (DuoCore-FS-slow) outperforms $\pi_{0}$~\cite{black2410-pi0} in out-of-distribution settings, achieving 70\% vs. 20\% success on “pick up the cup”. This gain is partly attributable to the slow system’s token-by-token (autoregressive) reasoning. However, DuoCore-FS-slow remains weaker on fine-grained subtasks such as “Grab the scoop” and “Scoop the popcorn,” consistent with Table~\ref{tab:result_in_distribution}. The full DuoCore-FS system combines the strengths of DuoCore-FS-slow and $\pi_{0}$~\cite{black2410-pi0}: it improves generalization while retaining $\pi_{0}$-level accuracy on fine-grained tasks.

\begin{table}[htbp]
\caption{\label{tab:result_out_distribution}\textbf{Results out-of-distribution.}}
\centering
\resizebox{\linewidth}{!}{%
\begin{tabular}{lccccc}
\toprule
Method & Pick up the cup & Grab the scoop & Scoop the popcorn &  Place the cup on table & Overall S.R. $\uparrow$ \\
\midrule
$\pi_{0}$\cite{black2410-pi0} & 20\% (2/10) & 50\% (1/2) & 100\% (1/1) & 100\% (1/1) & 10\% (1/10) \\
DuoCore-FS-slow & 70\% (7/10) & 57.1\% (4/7) & 25\% (1/4) & 100\% (1/1) & 10\% (1/10) \\
DuoCore-FS & 50\% (5/10) & 100\% (5/5) & 100\% (5/5) & 100\% (5/5) & 50\%(5/10) \\
\bottomrule
\end{tabular}%
}
\end{table}

\begin{figure}
\centering
\includegraphics[width=1.0\linewidth]{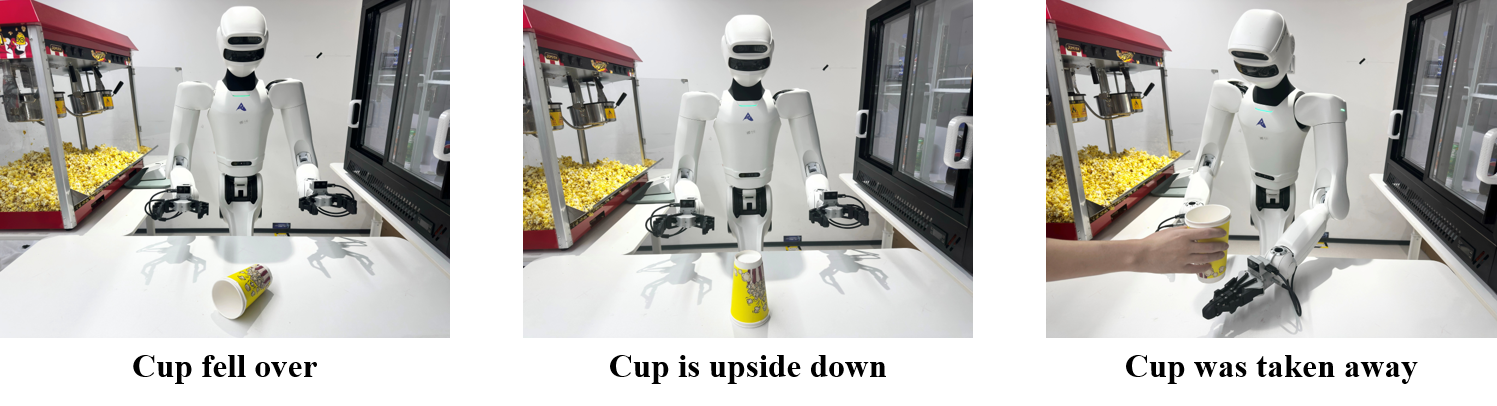}
\caption{\label{fig:anomaly case}\textbf{Illustration of anomaly cases}}
\end{figure}

\noindent\textbf{Results under Anomalous Scenarios.} 
To assess robustness in realistic deployment conditions, we included several anomalous scenarios commonly observed in commercial environments—such as “Cup fell over”, “Cup is upside down”, and “Cup was taken away”, as shown in Figure~\ref{fig:anomaly case}.
In these cases, the robot is expected to return to its home pose and remain idle until the cup is properly repositioned. Table~\ref{tab:result_anomaly_detection} reports performance under these anomalous scenarios. DuoCore-FS achieves a slightly higher detection and recovery success rate than $\pi_{0}$\cite{black2410-pi0}.

\begin{table}[htbp]
\caption{\label{tab:result_anomaly_detection}\textbf{Performance under anomalous scenarios}}
\centering
\resizebox{\linewidth}{!}{%
\begin{tabular}{lcccc}
\toprule
Method & Cup fell over & Cup is upside down & Cup was taken away & Overall S.R. $\uparrow$\\
\midrule
$\pi_{0}$\cite{black2410-pi0} & 87.5\% (7/8) & 87.5\% (7/8) & 100\% (8/8) & 91.7\% (22/24)\\
DuoCore-FS & 100\% (8/8) & 87.5\% (7/8) & 100\% (8/8) & 95.8\% (23/24)\\
\bottomrule
\end{tabular}%
}
\end{table}

\noindent\textbf{Results of Language Following.} 
We evaluate language-following ability by measuring the success rate of executing the instructed behavior under identical visual observations. Experiments are conducted on the popcorn-scooping and beverage-cabinet–closing tasks. Since their initial states were matched during data collection, we can assess language following by ensuring that the two tasks share identical image observations at test time. 

As Table~\ref{tab:result_language_following} shows, DuoCore-FS demonstrates substantially stronger language-following ability than $\pi_{0}$~\cite{black2410-pi0}, achieving a 42.9\% success rate compared to 14.3\%. We attribute this improved language-following performance to the slow system’s autoregressive training objective, which encourages more consistent token-by-token reasoning over instructions and observations. Despite this improvement, the success rate remains far from perfect, largely due to the extreme data imbalance: the beverage-cabinet closing task constitutes only 1\% of the training set, and no imbalance-mitigation strategies were applied. A typical failure case is instruction confusion, where the model executes “pick up the cup” when asked to “close the beverage cabinet door.”

\begin{figure}
\centering
\includegraphics[width=1.0\linewidth]{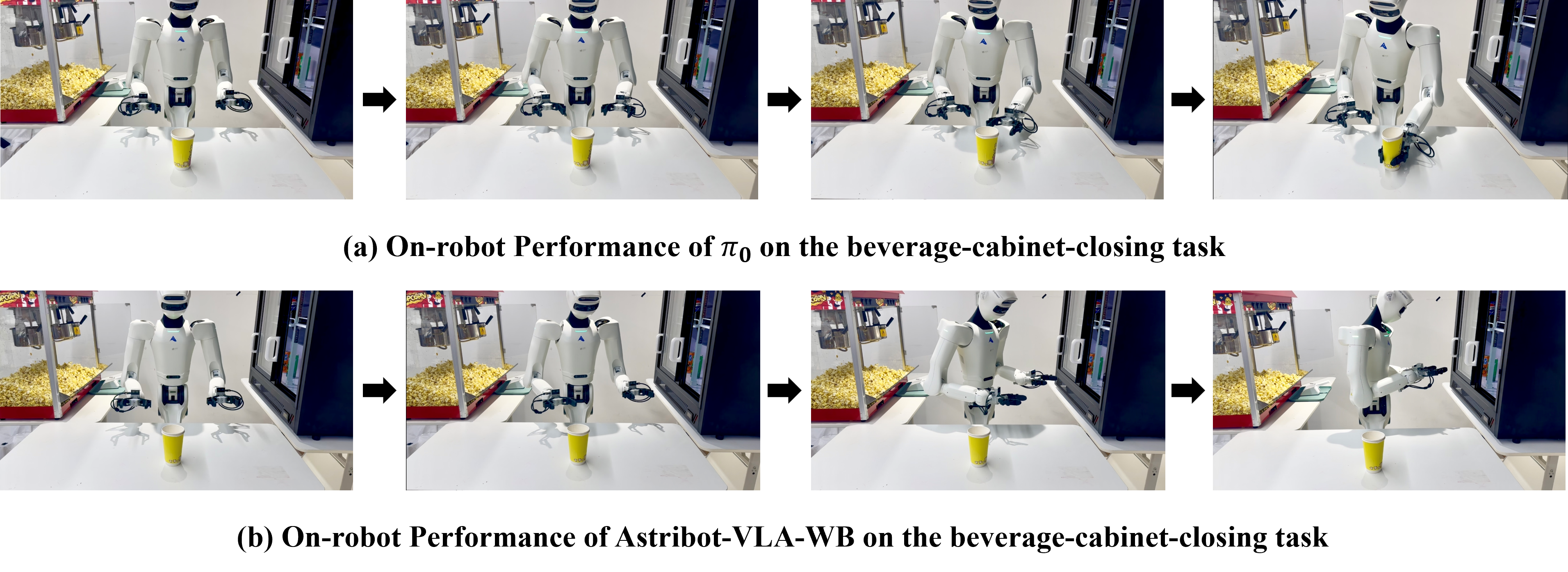}
\caption{\label{fig:language_follow}\textbf{On-robot performance comparison of $\pi_{0}$\cite{black2410-pi0} vs. DuoCore-FS on the beverage-cabinet-closing task .} When instructed to close the door, $\pi_{0}$\cite{black2410-pi0} consistently ignores the command and attempts to pick up the cup, as it assumes it should perform the popcorn-scooping task. In contrast, DuoCore-FS follows the instruction and closes the door.}
\end{figure}

\begin{table}[htbp]
\caption{\label{tab:result_language_following}\textbf{Results of language following} }
\centering
\scalebox{0.8}{
\begin{tabular}{lc}
\toprule
Method & Close the beverage cabinet door\\
\midrule
$\pi_{0}$\cite{black2410-pi0} & 14.3\% (1/7) \\
DuoCore-FS & 42.9\% (3/7) \\
\bottomrule
\end{tabular}%
}
\end{table}

\noindent\textbf{Ablation Study of Action Tokenizer.} 
We evaluate the performance of two tokenizers used in our slow system, FAST\cite{pertsch2025fast} tokenizer and our RVQ‑VAE tokenizer on the popcorn‑scooping task. We find that naïvely tokenizing full-body actions with the fast tokenizer causes a combinatorial explosion, producing sequences that can span thousands of action tokens. To mitigate this, we tokenize the robot’s three $SO(3)$ components separately and predict action tokens sequentially for the left hand, right hand, and torso $SO(3)$. As Table.\ref{tab:ab_study_action_tokenizer} shows, when applying the FAST method~\cite{pertsch2025fast}, the resulting action tokens are on average over 2× longer than those produced by our RVQ-VAE, with the maximum length exceeding exceeds 5×. This substantially constrains inference speed, limiting it to 0.95 Hz. Furthermore, sequential prediction of the left hand, right hand, and torso action tokens induces compounding errors. In deployment, our slow system with FAST\cite{pertsch2025fast} frequently generates incorrect tokens, leading to erratic motions and no successful task executions.

\begin{table}[htb]
\caption{\label{tab:ab_study_action_tokenizer}Result of action tokenizer}
\centering
\scalebox{0.7}{
\begin{tabular}{lcccc}
\toprule
Tokenizer & Avg. token length & Maximum token length & Sucess rate & Inference frequency (Hz) \\
\midrule
FAST\cite{pertsch2025fast} & 81.12 & 205 & 0\% & 0.95 \\
RVQ-VAE & 36 & 36 & 83.3\% & 3.27 \\
\bottomrule
\end{tabular}%
}
\end{table}

\section{Conclusion, Limitations, and Future work}
We introduced DuoCore-FS, a fast–slow VLA framework that enables high-frequency whole-body action generation while leveraging the rich semantic reasoning of a large VLM. The design allows the system to sustain 30 Hz action-chunk generation with a 3B-scale VLM, over three times the inference frequency of prior VLA models \cite{black2410-pi0, intelligence2025pi_} with comparable parameter sizes, while remaining fully end-to-end trainable. Evaluations on real-world whole-body manipulation tasks show that our approach maintains task success rates while substantially improving high-frequency action generation compared to synchronous fast–slow VLA baselines.

While effective, DuoCore-FS still presents several limitations. Future work may explore:

\begin{itemize}
  \item DuoCore-FS is a highly extensible framework. The slow pathway can be enhanced through multi-task foundation model pretraining for manipulation (e.g., subtask prediction, object grounding, affordance reasoning), augmented with more diverse real-world manipulation data, and through chain-of-thought reasoning at inference time. 
  \item Accelerating the fast pathway by reducing the number of flow-matching steps or moving toward single-step generation, enabling higher action-chunk generation frequencies.
  \item Designing more effective and sophisticated bridging mechanisms between the slow and fast pathways to improve task success rates and the real-time performance of whole-body manipulation.
  \item Incorporating high-frequency force and tactile signals could improve responsiveness in contact-rich tasks, taking advantage of the fast–slow architecture to handle multi-modal inputs at their respective temporal frequencies. 
  \item Evaluating the framework on a broader set of real-world tasks, including dynamic scenarios and long-horizon manipulation tasks, will be important to fully validate its robustness and generalization.
\end{itemize}

\section{Contributions and Acknowledgments}
\begin{itemize}
  \item \textbf{Core Contributors:} Teqiang Zou, Hongliang Zeng
  \item \textbf{Contributors:} Yuxuan Nong, Yifan Li, Kehui Liu, Haotian Yang, Xinyang Ling, Xin Li
  \item \textbf{Project Lead:} Lianyang Ma
  \item \textbf{Acknowledgments: }We would like to thank Guang Gao for the insightful discussion on morphological mirror augmentation. We thank Jinbo Zuo, Zhaohui An, and Ruirui Zhang  for their help on robotic system development and maintenance.
\end{itemize}
\label{sec:contrib}

\printbibliography

\end{document}